\documentclass{article}
\usepackage{amsmath,epsfig}
\usepackage[preprint]{spconfa4}
\usepackage{xcolor}
\usepackage{cite}

\usepackage{rotating}
\usepackage{enumitem}
\usepackage{makecell}
\usepackage{multirow}
\usepackage{graphicx}
\usepackage[caption=false]{subfig}
\usepackage{amssymb}
\usepackage{pifont}
\newcommand{\cmark}{\ding{51}}%

\usepackage{url}
\makeatletter
\let\NAT@parse\undefined
\makeatother
\usepackage[breaklinks=true, colorlinks, bookmarks=false]{hyperref}

\let\OLDthebibliography\thebibliography
\renewcommand\thebibliography[1]{
\OLDthebibliography{#1}
\setlength{\parskip}{0pt}
\setlength{\itemsep}{0pt plus 0.3ex}
}

\begin{document}\sloppy

\def\x{{\mathbf x}}
\def\L{{\cal L}}

\title{CENet: Toward Concise and Efficient LiDAR Semantic Segmentation\\ for Autonomous Driving}

\name{Hui-Xian Cheng, Xian-Feng Han\thanks{\textsuperscript{*} Corresponding author. This research was supported by the National Natural Science Foundation of China (No. 62002299), and the Natural Science Foundation of Chongqing, China (No. cstc2020jcyj-msxmX0126), and the Fundamental Research Funds for the Central Universities (No. SWU120005)}\textsuperscript{*},   Guo-Qiang Xiao}
\address{College of Computer and Information Science, Southwest University, Chongqing, China 
	\\ \tt\small\{chenghuixian\}@email.swu.edu.cn \quad \{xianfenghan, gqxiao\}@swu.edu.cn}

\maketitle

\begin{abstract}

Accurate and fast scene understanding is one of the challenging task for autonomous driving, which requires to take full advantage of LiDAR point clouds for semantic segmentation. In this paper, we present a \textbf{concise} and \textbf{efficient} image-based semantic segmentation network, named \textbf{CENet}. In order to improve the descriptive power of learned features and reduce the computational as well as time complexity, our CENet integrates the convolution with larger kernel size instead of MLP, carefully-selected activation functions, and multiple auxiliary segmentation heads with corresponding loss functions into architecture. Quantitative and qualitative experiments conducted on publicly available benchmarks, SemanticKITTI and SemanticPOSS, demonstrate that our pipeline achieves much better mIoU and inference performance compared with state-of-the-art models. The code will be available at \url{https://github.com/huixiancheng/CENet}.

\end{abstract}
\section{Introduction}
\label{sec:intro}

\begin{figure}[t]
	\vspace{-4mm}
	\includegraphics[width=0.98\columnwidth, height=5.8cm]{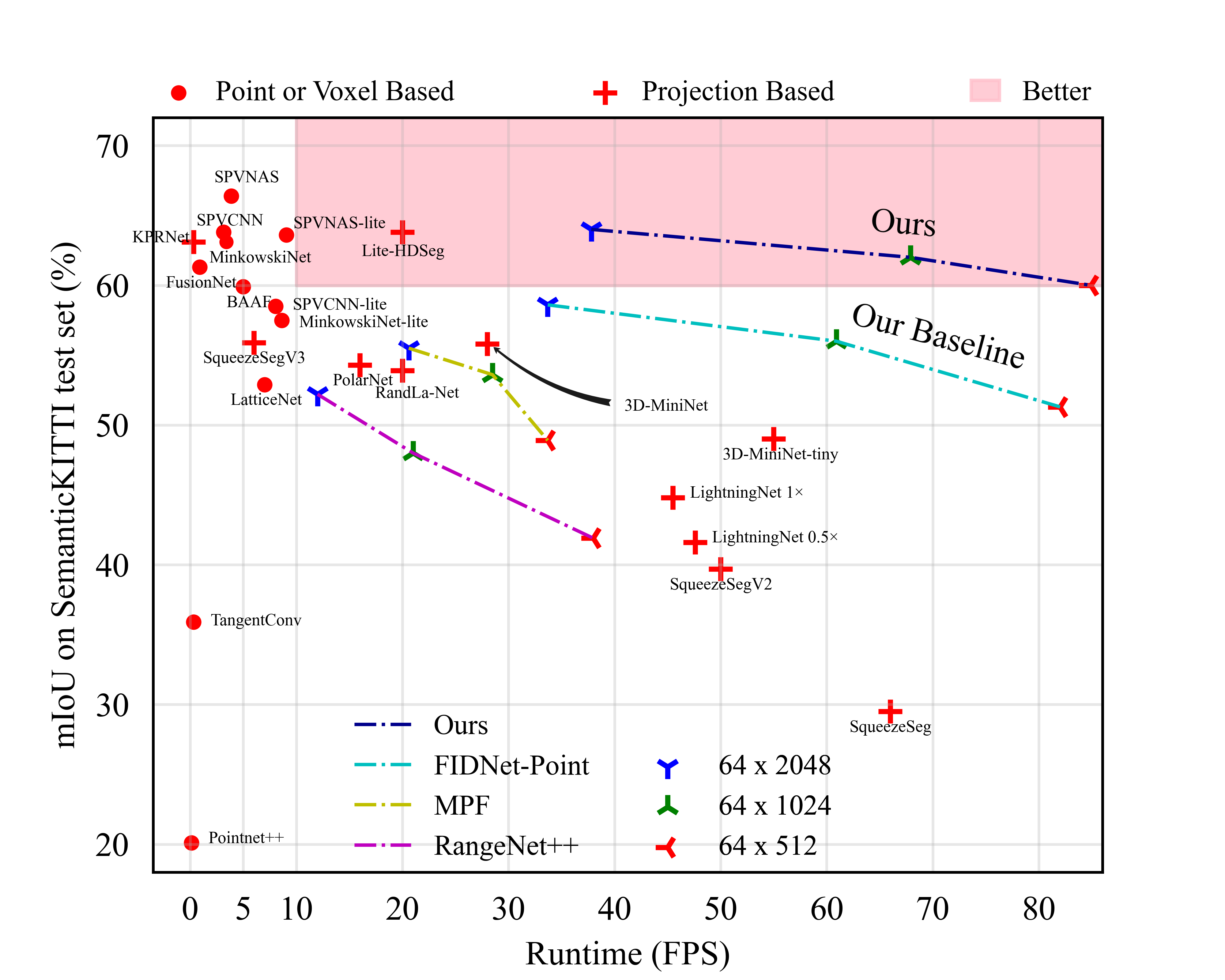}
	\vspace{-4mm}
	\caption{LIDAR semantic segmentation \textbf{accuracy} vs \textbf{speed} on SemanticKITTI test set \cite{behley2019semantickitti}. (The dotted lines represent the performance of the same method at different resolutions.) \textit{Best viewed in color and zoomed in for more detail.}}
	\label{fig:first}
	\vspace{-6mm}
\end{figure}

Recently, the rapid development of LiDAR sensors makes the 3D computer vision become an interesting research topic in applications, such as robotics and autonomous driving \cite{feng2020deep}, where accurate, real-time and robust environment perception and understanding is a challenging task. In order to achieve this goal, LiDAR becomes the most popular and widely used choice, since 1) compared with visual cameras, LiDAR is much more robust to varying lighting and weather conditions. 2) and the acquired 3D point clouds provide rich geometric information. Therefore, LiDAR-based semantic scene perception, especially 3D point cloud semantic segmentation, has received increasing attention, aiming to perform point-wise classification.

In the past few years, the emergence of datasets, such as SemanticKITTI~\cite{behley2019semantickitti} and SemanticPOSS~\cite{pan2020semanticposs}, benchmarks the LiDAR semantic segmentation. And they make it possible to apply deep learning techniques to this task. However, we cannot directly perform standard Convolutional Neural Networks (CNNs) on 3D point clouds due to their irregular and sparse structure. To address this problem, many recent new methods have been proposed, which can be classified into raw point-based~\cite{qi2017pointnet++, thomas2019kpconv, hu2020randla}, voxel-based~\cite{tang2020searching, zhou2020cylinder3d}, and range image-based~\cite{wu2018squeezeseg,milioto2019rangenet++}.     

Generally, point-based networks deal with raw 3D LiDAR point clouds directly, which can obtain much better performance with higher computational complexity ~\cite{thomas2019kpconv} \cite{hu2020randla}. Voxel-based methods project unstructured point clouds into regular grid cells, and allows the usage of 3D convolutional neural networks. Although the state-of-the-art accuracy can be achieved, the higher model complexity makes it difficult for these methods to obtain real-time inference speed. For example, the inference speeds of SPVCNN~\cite{tang2020searching} and Cylinder3D~\cite{zhou2020cylinder3d} on Tesla V100 are only 8.0 and 7.6 fps, respectively. In addition, range image-based approaches choose to represent the raw 3D point clouds as an ordered range image using spherical projection strategy, then the well-designed 2D CNNs can be utilized to carry out LiDAR semantic segmentation task. This kinds of methods actually can provide superior inference and accuracy performance, which drives more and more researchers to focus on this field \cite{fan2021rangedet}, though they inevitably suffer from information loss during projection \cite{guo2020deep}.

On the other hand, the current range image-based methods, such as KPRNet~\cite{kochanov2020kprnet} and Lite-HDSeg~\cite{Lite-HDSeg} typically have an extremely large number of parameters, and much lower inference speed or higher model complexity, which limits their applications in autonomous driving to some extent. In order to relieve this issue, FIDNet~\cite{zhao2021fidnet} presents a new range image-based LiDAR semantic segmentation network that keeps the solution as simple as possible while maintaining good performance.

Nevertheless, the performance of FIDNet is not comparable to the current state-of-the-art methods~\cite{qiu2021semantic, tang2020searching, Lite-HDSeg}. So based on the fundamental idea and performance of FIDNet, we attempt to rethink its design choices, and propose a concise and efficient LiDAR semantic segmentation model, termed CENet. Quantitatively experimental results demonstrate that our network performance can outperform current state-of-the-art methods with no increase in the number of effective parameters, and has a much higher inference speed (as shown in Fig.~\ref{fig:first}). Specifically, the main contributions of this paper are as follows:

\begin{itemize}[leftmargin=*]
\vspace{-1.5ex}
\item We present a newly-designed LiDAR point cloud segmentation architecture, named CENet, which can improve the inference speed at the cost of no parameters increment.
\vspace{-1.5ex}
\item In order to improve the nonlinear capability of the network, we use SiLU and Hardswish to adjust the activation functions.
\vspace{-1.5ex}
\item By introducing multiple auxiliary segmentation heads, we significantly improve the learning power of our network without introducing additional inference parameters.
\vspace{-1.5ex}
\item We conduct comprehensive experiments on the publicly available datasets, SemanticKITTI and SemanticPOSS. The results show that our method achieves state-of-the-art performance.
\end{itemize}

\section{Related work}
\label{sec:related work}
\textbf{Point-based methods} work directly on the raw point clouds. PointNet~\cite{qi2017pointnet} and PointNet++\cite{qi2017pointnet++} are pioneering studies to use shared MLPs to learn the properties of each point, which inspire the appearance of a series of point-based networks. KPConv~\cite{thomas2019kpconv} develops deformable convolutions that can use arbitrary number of kernel points to learn local representations. RandLA-Net~\cite{hu2020randla} adopts a random sampling strategy to considerably improve the efficiency of point cloud processing and uses local feature aggregation to reduce the information loss caused by random operations.  BAAF~\cite{qiu2021semantic} makes full use of the geometric and semantic features of points to obtain more accurate semantic segmentation by using bilateral structures and adaptive fusion methods. 

\noindent\textbf{Voxel-based methods} first discretize the point clouds into 3D voxel representations, then predict semantic labels for these voxels using 3D CNN frameworks. Minkowski~\cite{graham20183d} chose to use sparse convolution instead of standard 3D convolution to reduce the computational cost. SPVNAS~\cite{tang2020searching} exploits neural structure search (NAS) to further improve the network's performance. 

\noindent\textbf{Image-based methods} project LiDAR point clouds onto 2D multimodal images and then apply the well-designed 2D CNNs for semantic segmentation. SqueeeSeg~\cite{wu2018squeezeseg} and SqueezeSegV2~\cite{wu2019squeezesegv2} use the lightweight model SqueezeNet and CRF for segmentation. RangeNet++~\cite{milioto2019rangenet++} integrates Darknet into SqueezeSeg and proposes an efficient KNN post-processing method to predict labels for point. SqueezeSegV3~\cite{xu2020squeezesegv3} proposes Spatially-Adaptive Convolution (SAC) with different filters depending on the location of the input image. KPRNet~\cite{kochanov2020kprnet} achieves promising results by using powerful as backbone together with KPConv as segmentation head. Lite-HDSeg~\cite{Lite-HDSeg} achieves state-of-the-art performance by introducing three different modules, Inception-like Context Module, Multi-class Spatial Propagation Network, and a boundary loss. Although it gets a high inference speed, the higher number of parameters and model complexity make it less suitable for autonomous driving applications.

\section{METHODOLOGY}
\label{sec:method}
\begin{figure*}[t]
\centering
\includegraphics[width=\linewidth]{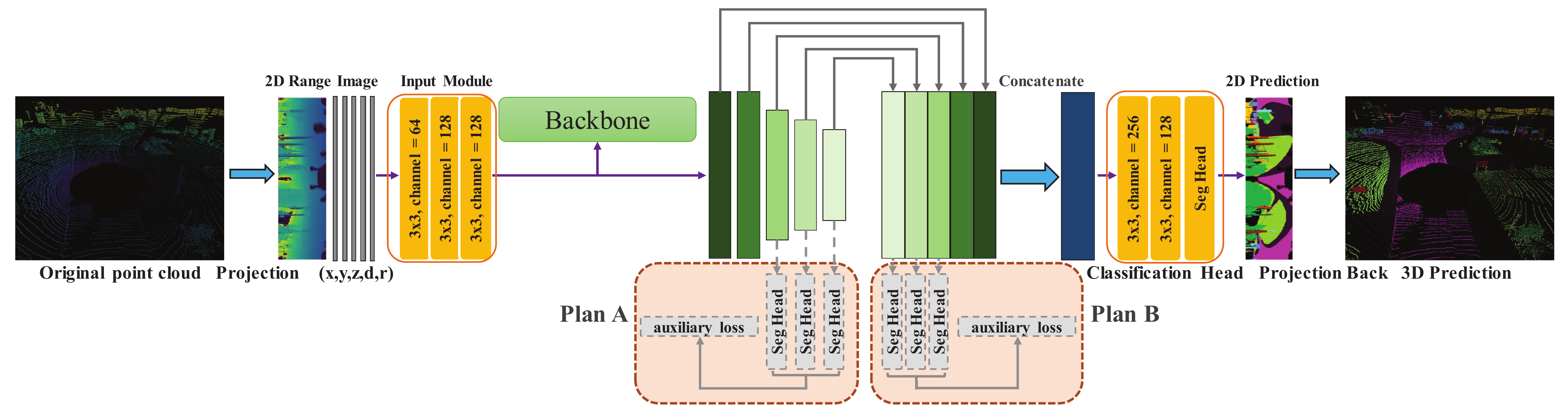}
\vspace{-5mm}
\caption{The overall architecture of our pipeline. The backbone network can be any feature extraction structure such as BasicBlock in ResNet used in this paper. Plan A and Plan B are two different designs of auxiliary loss. As shown in the dotted line, it can be removed during inference thus not influencing the inference speed.}
\label{fig:framework}
\vspace{-6mm}
\end{figure*}

In order to take full advantage of well-optimized conventional convolution functions for real-time LiDAR point cloud segmentation, we use the spherical projection approach to generate 2D multimodal range images by projecting the LiDAR point cloud into the spherical coordinate system, which is formulated as, 
\vspace{-1.5ex}
\begin{equation}
\label{proj}
\left(
\begin{matrix}
	u \\
	v
\end{matrix}
\right) =
\left(
\begin{matrix}
	\frac{1}{2}\left [ 1-arctan\left ( y,x \right ) \pi^{-1} \right ]W \\
	\left [ 1 - \left (  arcsin\left ( z, d^{-1} \right ) + f_{u} \right )  \frac{1}{f_{u} + f_{d} }\right ]H
\end{matrix}
\right)
\vspace{-1.5ex}
\end{equation}

where $f = f_{u}$ + $f_{d}$ refers to the sensor's vertical field-of-view. The depth $d$ of each point is calculated as $d = \sqrt{x^{2} + y^{2} + z^{2}}$.
The final result is a projected range image of size $(H, W, 5)$, where each pixel contains 5 channels $(x, y, z, d, r)$ and $r$ is the intensity information of points. Based on this transformation, the point cloud segmentation problem is turned into an image segmentation task.

\subsection{Network Architecture}
\label{sec:network arch}
Our CENet architecture is shown in Fig~\ref{fig:framework}. We will detailedly introduce the core components in the following subsections.

\noindent\textbf{Input Module, Classification Head and Activation Function.} The range image is a special two-dimensional image with five channels, where each location actually can be considered as a point representation. Therefore, FIDNet ~\cite{zhao2021fidnet} uses $1 \times 1$ conv layers in \textbf{Input Module} and \textbf{Classification Head} to process the input features, which, the authors think, can achieve the similar effect as MLP used in PointNet for point feature learning. However, it is more reasonable to use $3\times3$ conv layers instead of $1\times1$, due to the following reasons. 1) Unlike disordered and unstructured point cloud, the generated range images usually have structure features, which makes $3\times3$ conv much more suitable than MLP. 2) For $1\times1$ conv, the lower number of parameters and computational cost does not mean faster inference speed. As shown in Table~\ref{tab:conv} from RepVGG~\cite{ding2021repvgg}, it can be concluded that under the same condition, the computational density (theoretical operations divided by time usage) of $3\times3$ conv can be up to $4\times$ that of $1\times1$ conv on a 1080Ti GPU. Therefore, we attempt to conduct a simple substitution to improve the inference speed, which will be verified in \textbf{ Sec~\ref{sec:abla} Ablation Studies}. In addition, inspired by YoloV5 and MobileNetV3, we can see that the usage of a stronger nonlinear activation function can contribute to the improvement of network's expressive power without increasing the parameters. Hence, we adopt SiLU and Hardswish activation functions in our model. Experiments in \textbf{Sec~\ref{sec:abla}} also show that both functions can enhance our model's performance with little influence on inference speed.

\begin{table} [t]\footnotesize 
\centering
\caption{Inference speed test with varying kernel size on NVIDIA 1080Ti. The batch size = 32, input channels = output channels = 2048, resolution = 56$\times$56, stride = 1.}
\label{tab:conv}
	\begin{tabular}{lccc} 
		\hline
		\begin{tabular}[c]{@{}l@{}}Kernel \\ size\end{tabular} & \begin{tabular}[c]{@{}c@{}}Theoretical \\ FLOPs (B)\end{tabular} & \begin{tabular}[c]{@{}c@{}}Time\\usage (ms)\end{tabular} & \begin{tabular}[c]{@{}c@{}}Theoretical \\ TFLOPS\end{tabular} \\ 
		\hline
		$1\times1$ & 420.9 & 84.5 & 9.96 \\
		$3\times3$ & 3788.1 & 198.8 & \textbf{38.10} \\
		\hline
	\end{tabular}
\vspace{-6mm}
\end{table}

\begin{table*}[ht!] 
\begin{center}
	\caption{The performance comparison on SemanticKITTI test set. }
	\label{tab:kitti}
	\resizebox{0.9\textwidth}{!}{
		\begin{tabular}{c|c|c|c|c|ccccccccccccccccccc}
			Category  & Methods & Size & \rotatebox{90}{\textbf{FPS~(Hz)}} & \rotatebox{90}{\textbf{mean-IoU}} & \rotatebox{90}{car}& \rotatebox{90}{bicycle}& \rotatebox{90}{motorcycle}& \rotatebox{90}{truck}& \rotatebox{90}{other-vehicle}& \rotatebox{90}{person}& \rotatebox{90}{bicyclist}& \rotatebox{90}{motorcyclist}& \rotatebox{90}{road}& \rotatebox{90}{parking}& \rotatebox{90}{sidewalk}& \rotatebox{90}{other-ground}& \rotatebox{90}{building}& \rotatebox{90}{fence}& \rotatebox{90}{vegetation}& \rotatebox{90}{trunk}& \rotatebox{90}{terrain}& \rotatebox{90}{pole}& \rotatebox{90}{traffic-sign} \\ 
			\hline
			\hline
			\multirow{4}{*}{Point-based} &PointNet++~\cite{qi2017pointnet++} & \multirow{4}{*}{50K pts} & 0.1 & 20.1 & 53.7 & 1.9 & 0.2 & 0.9 & 0.2 & 0.9 & 1.0 & 0.0 & 72.0 & 18.7 & 41.8 & 5.6 & 62.3 & 16.9 & 46.5 & 13.8 & 30.0 & 6.0 & 8.9 \\ 
			&RandLa-Net~\cite{hu2020randla} &  & 20 & 53.9 & 94.2 & 26.0 & 25.8 & 40.1 & 38.9 & 49.2 & 48.2 & 7.2 & 90.7 & 60.3 & 73.7 & 20.4 & 86.9 & 56.3 & 81.4 & 61.3 & 66.8 & 49.2 & 47.7 \\
			&KPConv~\cite{thomas2019kpconv} &  & $-$ & 58.8 & \textbf{96.0} & 30.2 & \textbf{42.5} & 33.4 & 44.3 & \textbf{61.5} & \textbf{61.6} & 11.8 & 88.8 & 61.3 & 72.7 & \textbf{31.6} & \textbf{90.5} & \textbf{64.2} & \textbf{84.8} & \textbf{69.2} & \textbf{69.1} & \textbf{56.4} & 47.4 \\ 
			&BAAF~\cite{qiu2021semantic} &  &5 & 59.9 & 95.4 & \textbf{31.8} & 35.5 & \textbf{48.7} & \textbf{46.7} & 49.5 & 55.7 & \textbf{53.0} & \textbf{90.9} & \textbf{62.2} & \textbf{74.4} & 23.6 & 89.8 & 60.8 & 82.7 & 63.4 & 67.9 & 53.7 & \textbf{52.0}\\ 
			\hline
			\multirow{4}{*}{Voxel-based}&MinkowskiNet-lite~\cite{graham20183d} & \multirow{4}{*}{Voxel} & 8.6
			& 57.5 & $-$ & $-$ & $-$ & $-$ & $-$& $-$& $-$& $-$& $-$& $-$& $-$& $-$& $-$& $-$& $-$& $-$& $-$& $-$& $-$ \\
			&MinkowskiNet~\cite{graham20183d} & & 3.4
			& 63.1 & $-$ & $-$ & $-$ & $-$ & $-$& $-$& $-$& $-$& $-$& $-$& $-$& $-$& $-$& $-$& $-$& $-$& $-$& $-$& $-$ \\
			&SPVCNN-lite~\cite{tang2020searching} & & 8.1
			& 58.5 & $-$ & $-$ & $-$ & $-$ & $-$& $-$& $-$& $-$& $-$& $-$& $-$& $-$& $-$& $-$& $-$& $-$& $-$& $-$& $-$ \\
			&SPVCNN~\cite{tang2020searching} & & 3.2
			& 63.8 & $-$ & $-$ & $-$ & $-$ & $-$& $-$& $-$& $-$& $-$& $-$& $-$& $-$& $-$& $-$& $-$& $-$& $-$& $-$& $-$ \\
			\hline
			\multirow{19}{*}{Image-based} & SqueezeSeg-CRF~\cite{wu2018squeezeseg} & \multirow{6}{*}{$64\times 2048$} & 55 & 30.8 & 68.3 & 18.1 & 5.1 & 4.1 & 4.8 & 16.5 & 17.3 & 1.2 & 84.9 & 28.4 & 54.7 & 4.6 & 61.5 & 29.2 & 59.6 & 25.5 & 54.7 & 11.2 & 36.3 \\
			& SqueezeSegV2-CRF~\cite{wu2019squeezesegv2} &  & 40 & 39.6 & 82.7 & 21.0 & 22.6 & 14.5 & 15.9 & 20.2 & 24.3 & 2.9 & 88.5 & 42.4 & 65.5 & 18.7 & 73.8 & 41.0 & 68.5 & 36.9 & 58.9 & 12.9 & 41.0 \\
			& SqueezeSegV3~\cite{xu2020squeezesegv3} &  & 6 & 55.9 & 92.5 & 38.7 & 36.5 & 29.6 & 33.0 & 45.6 & 46.2 & 20.1 & 91.7 & 63.4 & 74.8 & 26.4 & 89.0 & 59.4 & 82.0 & 58.7 & 65.4 & 49.6 & 58.9 \\
			& SalsaNext~\cite{cortinhal2020salsanext} &  & 24 & 59.5 & 91.9 & 48.3 & 38.6 & \textbf{38.9} & 31.9 & 60.2 & 59.0 & 19.4 & 91.7 & 63.7 & 75.8 & 29.1 & 90.2 & 64.2 & 81.8 & 63.6 & 66.5 & 54.3 & 62.1 \\
			& Lite-HDSeg~\cite{zhao2021fidnet} &  & 20 & 63.8 & 92.3 & 40.0 & \textbf{55.4} & 37.7 & 39.6 & 59.2 & \textbf{71.6} & \textbf{54.3} & 93.0 & 68.2 & 78.3 & 29.3 & 91.5 & 65.0 & 78.2 & 65.8 & 65.1 & \textbf{59.5} & \textbf{67.7} \\
			& KPRNet~\cite{zhao2021fidnet} &  & 0.3 & 63.1 & \textbf{95.5}& 54.1 & 47.9 & 23.6 & \textbf{42.6} & \textbf{65.9} & 65.0 & 16.5 & \textbf{93.2} & \textbf{73.9} & \textbf{80.6} & \textbf{30.2} & \textbf{91.7} & \textbf{68.4} & \textbf{85.7} & \textbf{69.8} & \textbf{71.2} & 58.7 & 64.1 \\
			\cline{2-24}
			& RangeNet++ \cite{milioto2019rangenet++} & \multirow{4}{*}{$64\times 512$} & 38.5 & 41.9 & 87.4 & 26.2 & 26.5 & 18.6 & 15.6 & 31.8 & 33.6 & 4.0 & \textbf{91.4} & 57.0 & 74.0 & 26.4 & 81.9 & 52.3 & 77.6 & 48.4 & 63.6 & 36.0 & 50.0 \\ 
			& MPF \cite{alnaggar2021multi}  & & 33.7 & 48.9 & 91.1 & 22.0 & 19.7 & 18.8 & 16.5 & 30.0 & 36.2 & 4.2 & 91.1 & 61.9 & 74.1 & 29.4 & 86.7 & 56.2 & \textbf{82.3} & 51.6 & \textbf{68.9} & 38.6 & 49.8\\ 	
			& FIDNet-Point~\cite{zhao2021fidnet} &  & 82.0 & 51.3 & 90.4 & 28.6 & 30.9 & 34.3 & 27.0 & 43.9 & 48.9 & 16.8 & 90.1 & 58.7 & 71.4 & 19.9 & 84.2 & 51.2 & 78.2 & 51.9 & 64.5 & 32.7 & 50.3 \\		
			& \textbf{Ours} & & \textbf{84.9} & \textbf{60.7} & \textbf{92.1} & \textbf{45.4} & \textbf{42.9} & \textbf{43.9} & \textbf{46.8} & \textbf{56.4} & \textbf{63.8} & \textbf{29.7} & 91.3 & \textbf{66.0} & \textbf{75.3} & \textbf{31.1} & \textbf{88.9} & \textbf{60.4} & 81.9 & \textbf{60.5} & 67.6 & \textbf{49.5} & \textbf{59.1} \\		
			\cline{2-24}					
			& RangeNet++ \cite{milioto2019rangenet++} & \multirow{4}{*}{$64\times 1024$} & 23.3 & 48.0 & 90.3 & 20.6 & 27.1 & 25.2 & 17.6 & 29.6 & 34.2 & 7.1 & 90.4 & 52.3 & 72.7 & 22.8 & 83.9 & 53.3 & 77.7 & 52.5 & 63.7 & 43.8 & 47.2 \\ 
			& MPF \cite{alnaggar2021multi} & & 28.5 & 53.6 & 92.7 & 28.2 & 30.5 & 26.9 & 25.2 & 42.5 & 45.5 & 9.5 & 90.5 & 64.7 & \textbf{74.3} & \textbf{32.0} & 88.3 & 59.0 & \textbf{83.4} & 56.6 & \textbf{69.8} & 46.0 & 54.9\\ 	
			& FIDNet-Point~\cite{zhao2021fidnet} & & 60.9 & 56.0 & 92.4 & 44.0 & 41.5 & 33.2 & 30.8 & 57.9 & 52.6 & 18.0 & \textbf{91.0} & 61.2 & 73.8 & 12.6 & 88.2 & 57.9 & 80.8 & 59.5 & 65.1 & 45.3 & 58.4 \\
			& \textbf{Ours} & & \textbf{67.9} & \textbf{62.3} & \textbf{93.0} & \textbf{50.5} & \textbf{47.6} & \textbf{41.7} & \textbf{43.4} & \textbf{64.5} & \textbf{65.2} & \textbf{32.5} & 90.5 & \textbf{65.5} & 74.1 & 29.2 & \textbf{90.9} & \textbf{65.4} & 81.6 & \textbf{65.4} & 65.6 & \textbf{55.9} & \textbf{61.0} \\
			\cline{2-24}
			& RangeNet++ \cite{milioto2019rangenet++} & \multirow{4}{*}{$64\times 2048$} & 12.8 & 52.2  & 91.4 & 25.7 & 34.4 & 25.7 & 23.0 & 38.3 & 38.8 & 4.8 & \textbf{91.8} & \textbf{65.0} & \textbf{75.2} & 27.8 & 87.4 & 58.6 & 80.5 & 55.1 & 64.6 & 47.9 & 55.9\\
			& MPF \cite{alnaggar2021multi} &  & 20.6 & 55.5 & \textbf{93.4} & 30.2 & 38.3 & 26.1 & 28.5 & 48.1 & 46.1 & 18.1 & 90.6 & 62.3 & 74.5 & 30.6 & 88.5 & 59.7 & 83.5 & 59.7 & 69.2 & 49.7 & 58.1 \\	
			& FIDNet-Point~\cite{zhao2021fidnet} &  & 33.7 & 58.6 & 93.0 & 45.7 & 42.0 & 27.9 & 32.6 & 62.6 & 58.1 & 30.5 & 90.8 & 58.3 & 74.9 & 20.1 & 88.5 & 59.5 & 83.1 & 64.3 & 67.8 & 52.6 & 60.0 \\
			& \textbf{Ours} &  & \textbf{37.8} & \textbf{64.7} & 91.9 & \textbf{58.6} & \textbf{50.3} & \textbf{40.6} & \textbf{42.3} & \textbf{68.9} & \textbf{65.9} & \textbf{43.5} & 90.3 & 60.9 & 75.1 & \textbf{31.5} & \textbf{91.0} & \textbf{66.2} & \textbf{84.5} & \textbf{69.7} & \textbf{70.0} & \textbf{61.5} & \textbf{67.6} \\		
			\hline	
	\end{tabular}}
\end{center}
\vspace{-5mm}
\end{table*}

\noindent\textbf{Loss Function.} In order to solve problems \textbf{1)} Class imbalance, \textbf{2)} The problem of optimizing the intersection-over-union (IoU), \textbf{3)} Blurred segmentation boundaries, following  \cite{Lite-HDSeg}, we use three different loss functions, namely weighted cross-entropy loss $L_{wce}$, Lovász-Softmax loss $L_{ls}$ and boundary loss $L_{bd}$, to supervise our model.

For segmentation task, the boundary-blurring problems between different objects generally arise from upsampling and downsampling operations. To address this issue, we introduce a boundary loss function, which can be formally defined as, 

\vspace{-1ex}
\begin{equation}
L_{bd}(\hat{y}, y) = 1-\frac{2P^{c}R^{c}}{P^{c} + R^{c}}
\vspace{-1ex}
\end{equation}
Where $P^c$ and $R^c$ denote the precision and recall of the predicted boundary map $y_{pd}$ with respect to the ground truth $y_{gt}$ for class c. We give the definition of boundaries as,
\vspace{-1ex}
\begin{equation}
\left\{\begin{matrix}
	y_{gt}^{b}=pool(1-y_{gt},\theta _{0})-(1-y_{gt})
	\\ 
	y_{pd}^{b}=pool(1-y_{pd},\theta _{0})-(1-y_{pd})
\end{matrix}\right.
\vspace{-1ex}
\end{equation}
Here pool($\cdot$) refers to the max-pooling operation on a sliding window of size $\theta_0$. Finally, our total loss is the weighted combination of these three loss functions. The formulation is 
\vspace{-1ex}
\begin{equation}
\label{eq: loss}
L = \alpha L_{wce} + \beta L_{ls} + \gamma L_{bd}
\vspace{-1ex}
\end{equation}
where $\alpha$, $\beta$, and $\gamma$ are the corresponding weights and are set to 1.0, 1.5, 1.0 respectively. In addition, we set $\theta_0$ in $L_{bd}$ to be 3.

\noindent\textbf{Auxiliary Loss.} In FIDNet, the authors integrate bilinear upsampling methods into FID module to interpolate the low-resolution feature maps, which generates five point-wise feature tensors with the same resolution but encoding different levels of information.  Compared with conventional decoders used in other networks~\cite{cortinhal2020salsanext,Lite-HDSeg, kochanov2020kprnet}, FID is totally parameter-free and dramatically reduces the complexity and storage cost. However, this simple decoder make model's performance excessively depend on the low-dimensional and high-dimensional features. Additionally, unlike progressive upsampling decoder, the simple interpolation fusion decoding may results in feature maps at different scales not being fully aligned and decoded. To alleviate this problem, we introduce multiple auxiliary loss heads to refine the feature maps at different resolutions for learning capability improvement.

Specifically, we use the auxiliary segmentation head to predict the output of three feature maps with different resolutions, and compute the weighted loss together with the main loss to supervise our network to produce more semantic features. The final loss function can be defined as,
\vspace{-1ex}
\begin{equation}
\label{eq: final_loss}
L_{total}=L_{main} + \lambda \sum_{i=1}^{3}L(y_{i},\hat{y}_{i})
\vspace{-1ex}
\end{equation}
where $L_{main}$ is the main loss, $y_{i}$ is the semantic output obtained from stage $i$, and $\hat{y}_{i}$ represents the corresponding semantic label. $L(\cdot)$ is computed according to Equation~\ref{eq: loss}. As shown in Fig. \ref{fig:framework}, for Plan A,  $\hat{y}_{i}$ is obtained by downsampling the GT labels with corresponding rate. For Plan B, since all feature maps are upsampled to the final output size, the GT labels are their $\hat{y}_{i}$.

\begin{table}[t]\tiny
\vspace{-5mm}
\centering
\caption{Evaluation results on the SemanticPOSS test split.}
\label{tab:poss}
\setlength{\tabcolsep}{2.pt}
\renewcommand{\arraystretch}{1.} 
	\resizebox{\linewidth}{!}{
		\begin{tabular}{c|ccccccccccccc|c}
			\hline
			& 
			\rotatebox{90}{person} & 
			\rotatebox{90}{rider} & 
			\rotatebox{90}{car} & 
			\rotatebox{90}{truck} & 
			\rotatebox{90}{plants} & 
			\rotatebox{90}{traffic sign} & 
			\rotatebox{90}{pole} & 
			\rotatebox{90}{trashcan} & 
			\rotatebox{90}{building} & 
			\rotatebox{90}{cone/stone} & 
			\rotatebox{90}{fence} & 
			\rotatebox{90}{bike} &
			\rotatebox{90}{ground} & 
			\rotatebox{90}{mIoU}
			\\
			\hline
			SqueezeSeg \cite{wu2018squeezeseg} & 14.2 & 1.0 & 13.2 & 10.4 & 28.0 & 5.1 & 5.7 & 2.3 & 43.6 & 0.2 & 15.6 & 31.0 & 75.0 & 18.9\\
			SqueezeSeg + CRF \cite{wu2018squeezeseg} & 6.8 & 0.6 & 6.7 & 4.0 & 2.5 & 9.1 & 1.3 & 0.4 & 37.1 & 0.2 & 8.4 & 18.5 & 72.1 & 12.9\\
			SqueezeSegV2 \cite{wu2019squeezesegv2} & 48.0 & 9.4 & 48.5 & 11.3 & 50.1 & 6.7 & 6.2 & 14.8 & 60.4 & 5.2 & 22.1 & 36.1 & 71.3 & 30.0 \\
			SqueezeSegV2 + CRF \cite{wu2019squeezesegv2} & 43.9 & 7.1 & 47.9 & 18.4 & 40.9 & 4.8 & 2.8 & 7.4 & 57.5 & 0.6 & 12.0 & 35.3 & 71.3 & 26.9\\
			RangeNet53 \cite{milioto2019rangenet++} & 55.7 & 4.5 & 34.4 & 13.7 & 57.5 & 3.7 & 6.6 & 23.3 & 64.9 & 6.1 & 22.2 & 28.3 & 72.9 & 30.3\\
			RangeNet53 + KNN \cite{milioto2019rangenet++} & 57.3 & 4.6 & 35.0 & 14.1 & 58.3 & 3.9 & 6.9 & 24.1 & 66.1 & 6.6 & 23.4 & 28.6 & 73.5 & 30.9\\
			MINet \cite{li2020multi} & 61.8 & 12.0 & 63.3 & 22.2 & 68.1 & 16.3 & 29.3 & 28.5 & 74.6 & 25.9 & 31.7 & 44.5 & 76.4 & 42.7\\
			MINet + KNN \cite{li2020multi} & 62.4 & 12.1 & 63.8 & 22.3 & 68.6 & 16.7 & 30.1 & 28.9 & 75.1 & 28.6 & 32.2 & 44.9 & 76.3 & 43.2 \\
			FIDNet-Point \cite{zhao2021fidnet} & 71.6 & 22.7 & 71.7 & 22.9 & 67.7 & 21.8 & 27.5 & 15.8 & 72.7 & 31.3 & 40.4 & 50.3 & 79.5 & 45.8 \\
			FIDNet-Point + KNN \cite{zhao2021fidnet} & 72.2 & \textbf{23.1} & 72.7 & 23.0 & 68.0 & \textbf{22.2} & 28.6 & 16.3 & 73.1 & \textbf{34.0} & 40.9 & 50.3 & 79.1 & 46.4 \\
			\hline
			Ours & 74.9 & 21.8 & 77.0 & 25.3 & 72.0 & 18.0 & 30.9 & 46.9 & 75.9 & 26.1 & 47.5 & 51.7 & \textbf{80.7} & 49.9 \\
			Ours + KNN & \textbf{75.5} & 22.0 & \textbf{77.6} & \textbf{25.3} & \textbf{72.2} & 18.2 & \textbf{31.5} & \textbf{48.1} & \textbf{76.3} & 27.7 & \textbf{47.7} & \textbf{51.4} & 80.3 & \textbf{50.3} \\
			\hline
	\end{tabular}}
	\vspace{-5mm}
\end{table}

\section{Experiment}

\subsection{Dataset and Implementation details}
\noindent\textbf{SemanticKITTI} is a large-scale dataset for the task of point cloud segmentation of autonomous driving scenes. It contains 43,551 LiDAR scans from 22 sequences collected from a city in Germany, where sequences 00 to 10 (19,130 scans) are used for training, 11 to 21 (20,351 scans) for testing, and sequence 08 (4,071 scans) for validation. All sequences are labeled with dense point-wise annotations.

\noindent\textbf{SemanticPOSS} is a much smaller, sparser, and more challenging benchmark collected by Peking University. It consists of 2,988 different, complex LiDAR scenes, each with a large number of sparse dynamic instances (e.g. pedestrians and bicycles). SemanticPOSS is divided into 6 parts, where we use part 2 as test set, and others as training set. 

\noindent\textbf{Implementation details.}  We conduct all the experiments on a single NVIDIA RTX3060 and RTX3090 GPU. The Stochastic Gradient Descent (SGD) optimizer with momentum of 0.9 is use for network optimization. During training, we adopt random rotation, random point dropout, and addition of random noise to X, Y, Z values to perform data augmentation. The weight decay is set to 1$e^{-4}$. For SemanticsKITTI, we train network for 100 epochs with initial learning rate 1$e^{-2}$, which is dynamically adjusted by a cosine annealing scheduler. For SemantciPOSS, the network is trained for 3 cycles with 45 epochs, the min and max learning rate are set to 1$e^{-5}$ and 1$e^{-3}$, respectively. 

\subsection{Results and Discussion}

\begin{figure*}[ht]
	\vspace{-5mm}
	\centering
	\subfloat[Input Point Cloud]{%
		\includegraphics[width=0.24\linewidth,height=0.15\linewidth]{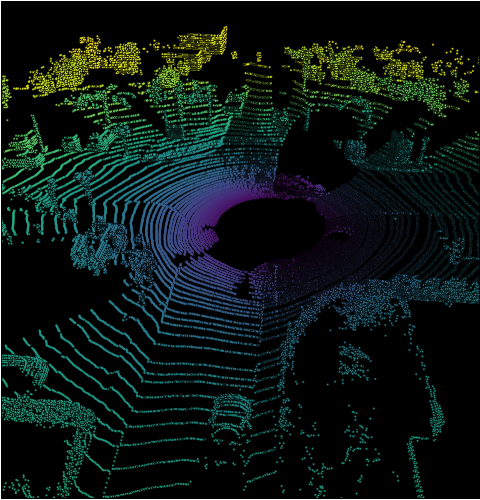}} \hspace{1mm}
	\subfloat[Ground Truth]{%
		\includegraphics[width=0.24\linewidth, height=0.15\linewidth]{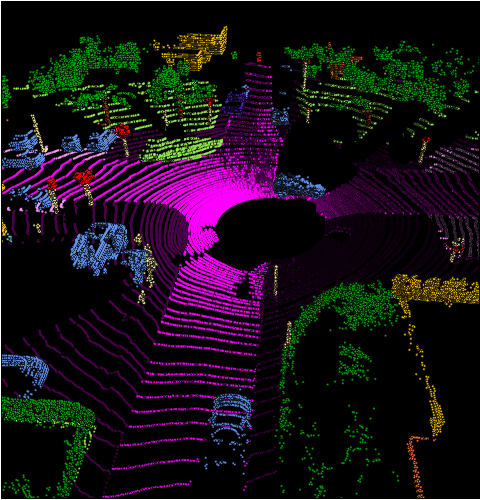}} \hspace{1mm}
	\subfloat[FIDNet-Point]{%
		\includegraphics[width=0.24\linewidth, height=0.15\linewidth]{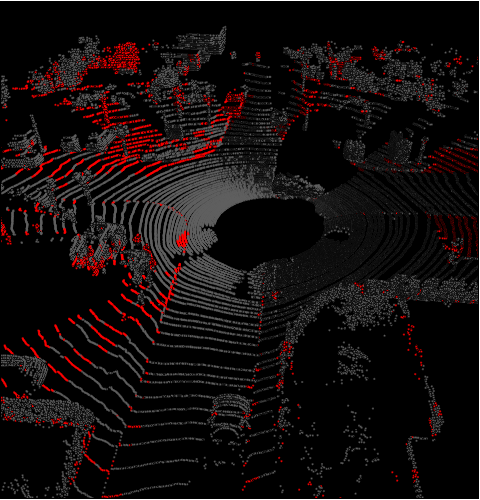}} \hspace{1mm}
	\subfloat[CENet(ours)]{%
		\includegraphics[width=0.24\linewidth, height=0.15\linewidth]{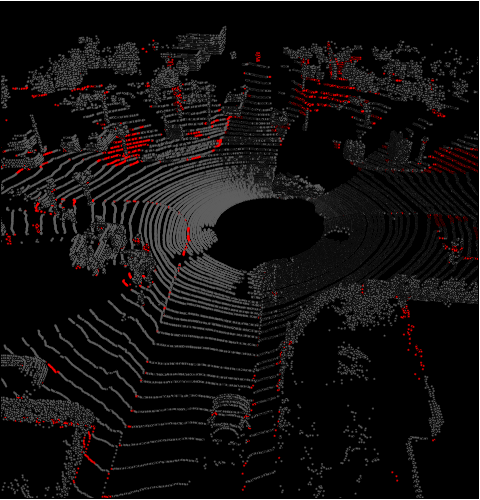}} \hspace{1mm}
	\vspace{-3mm}
	\caption{Qualitative analysis on SemanticKITTI validation set. Where \textbf{(a)} and \textbf{(b)} are input data of the LIDAR scan frame and corresponding segmentation ground truth, \textbf{(c)} and \textbf{(d)} are segmentation error maps in this scan frame for FIDNet and our method.~\textit{(With red indicating wrong prediction)}}
	\label{fig:view}
\end{figure*}

\noindent\textbf{Quantitative Results on SemanticKITTI:} Table~\ref{tab:kitti} shows the quantitative results of several state-of-the-art models on SemanticKITTI benchmark. It reports the input size, frames per second (FPS), mean IoU and class-wise IoU. The best results are highlighted in bold. From these results, it can be seen that for input of size $64\times2048$, our model achieves the state-of-the-art performance (64.7\%mIoU) compared to both point-based, voxel-based methods or image-based methods, while maintaining a higher FPS (37.8 FPS). For $64\times1024$ and $64\times512$ input, CENet obtain the superior results, 67.9 FPS, 62.3\%mIoU and 84.9FPS, 60.7\%mIoU, which outperforms the current methods by a large margin. It is worth noting that our CENet shows excellent performance when the input is $64\times512$, surpassing the performance of the baseline method FIDNet~\cite{zhao2021fidnet} and many other methods~\cite{xu2020squeezesegv3, cortinhal2020salsanext, alnaggar2021multi} under $64\times2048$ input.

\noindent\textbf{Quantitative Results on SemanticPOSS:} Table~\ref{tab:poss} shows the comparison of our proposed CENet with other related works. As can be reported from these results, all methods attain  a little worse results due to the differences in sensors and environments, as well as the samll-scale and sparse structure of features. Nevertheless, it is worth noting that our method outperforms all models not only in overall mIoU, but also almost class-wise mIoU. This further verify the effectiveness and efficiency of our network. Specifically, we achieve a significant $7.1\%$ and $3.9\%$ mIoU improvements over the  original best model MINet~\cite{li2020multi} and baseline method FIDNet~\cite{zhao2021fidnet}.

\noindent\textbf{Qualitative Result:} To better visualize the enhancement of our model over the baseline, we provide qualitative comparison examples in Fig~\ref{fig:view}. As can be seen from the results, our approach demonstrates a significant improvement over the baseline and is closer to the ground truth.

\subsection{Ablation Studies}

\begin{table}[t]
	\vspace{-5mm}
	\centering
	\caption{Ablation study evaluated on SemanticKITTI validation set.}
	\label{tab:ablation}
	\resizebox{\linewidth}{!}{
		\begin{tabular}{c|c|ccccc|c|c|c} 
			\hline
			Baseline & Row & KS & SiLU & H-swish & Plan A & Plan B & \textbf{mIoU} & \textbf{Params(M)} & \textbf{Latency(ms)}\\ 
			\hline\hline
			FIDNet &1 & &  &  & & & 55.4 & 6.053 & 11.956\\ 
			\hline\hline
			\multirow{12}{*}{Ours} &2& & &  &  &    & 58.9 & 6.053 & 11.956 \\ 
			\cline{2-10}
			&3& \cmark & & & && 59.2 & 6.774\label{line3} & 11.576 \\ 
			\cline{2-10}
			&4& & \cmark & &  &  & 60.0 & 6.053 & 11.981\\ 
			\cline{2-10}
			&5& & & \cmark &  &  & 60.5 & 6.053 & 12.039\\ 
			\cline{2-10}
			&6& \cmark & \cmark & &  &  & 59.5 &  6.774 & 11.776\\ 
			\cline{2-10}
			&7& \cmark & & \cmark &  &  & 60.6 &  6.774 & 11.562\\ 
			\cline{2-10}
			&8& & & & \cmark& & 62.5 & 6.061 & 11.956 \\  
			\cline{2-10}
			&9&  & & & &\cmark& 63.0 & 6.061 & 11.956 \\  
			\cline{2-10}
			&10& \cmark & & &  \cmark &  & 63.2 & 6.782 & 11.576\\ 
			\cline{2-10}
			&11& \cmark &  &  &  & \cmark & 64.3\label{line11} & 6.782 & 11.576\\ 
			\cline{2-10}
			&12& \cmark & & \cmark & & \cmark & 65.3 & 6.782 &11.562\\ 
			\hline
	\end{tabular}}
	\vspace{-5mm}
\end{table}

\begin{table} [t] \tiny
	\centering
	\caption{Impact of kenel size on time of model inference}
	\label{tab:speed}
	\resizebox{0.8\linewidth}{!}{
		\begin{tabular}{c|c|cc} 
			\hline
			Kernel~Size & Input Resolution & Model Latency(ms) & FPS \\ 
			\hline
			\multirow{3}{*}{$1\times1$} & 64 $\times$ 2048 & 29.347 & 33.7 \\
			& 64 $\times$ 1024 & 16.151 & 60.9 \\
			& 64 $\times$ 512 & 11.956 & 82.0 \\ 
			\hline
			\multirow{3}{*}{$3\times3$} & 64 $\times$ 2048 & 26.128 & 37.8 \\
			& 64 $\times$ 1024 & 14.578 & 67.4 \\
			& 64 $\times$ 512 & 11.576 & 84.8 \\
			\hline
	\end{tabular}}
	\vspace{-5mm}
\end{table}

\label{sec:abla}
To quantitatively analyze the effectiveness of different components, we conducted the following ablation experiments on the SemanticKITTI validation set. Here, we exploit a similar setup as SqueezeSegV3~\cite{xu2020squeezesegv3} for high efficient training and evaluation. The size of input range image is set to $64 \times 512$. And we choose to list the accuracy evaluated directly on the projected 2D image instead of original 3D points.

\noindent\textbf{Effects  of Module Components.} 
Table~\ref{tab:ablation} demonstrates the experimental results of different design choices of network, which uses the mIoU, traning parameters required by models, and inference time as measures. Results in \textbf{the first row} are obtained by using the official FIDNet code. \textbf{The second line} reports the performance achieved using our network, in which the required normal vector has been removed. It still get  3.5\% improvement over FIDNet. \textbf{The third row} validates the effectiveness when replacing the $1\times1$ convs with $3\times3$ convs. It can be concluded that although the $3\times3$ conv introduces more parameters than $1\times1$ conv, the former can further reduce the model's inference time, and slightly improve the performance by 0.3\%. This is mainly because large kernel size brings much larger perceptual field, and the modern computational library is highly optimized. Table~\ref{tab:speed} further reports the difference between $3\times3$ conv and $1\times1$ conv in model latency and total FPS for different input range image resolutions. Overall, $3\times3$ conv can improve the model inference speed significantly. \textbf{The fourth to seventh rows} state the performance using different activation functions. From these results, we can see that the introduction of stronger nonlinearities contributes to the descriptiveness improvement of our model at little cost of inference speed. Results in \textbf{the eighth to eleventh lines} show that 1) both auxiliary segmentation heads can significantly improve the performance of our CENet. Although the integration of the auxiliary heads introduces some additional training parameters and increases the training time, we can remove these auxiliary heads in the inference phase. Therefore, they have no effect on the network's latency. 2) The combination of $3\times3$ conv and auxiliary loss outperforms the effect in the original $1\times1$ conv with auxiliary loss. And 3) since Plan B module calculates the loss directly based on the upsampled feature maps, which serves to refinement of features at different stages, Plan B brings a much better performance than Plan A. 

\noindent\textbf{Effects  of $\lambda$ in auxiliary loss.} The hyperparameter $\lambda$ contained in Equation ~\ref{eq: final_loss} play an important role in optimizing the model's performance. Hence, we evaluate the effectiveness of $\lambda$. As illustrated in Table ~\ref{tab:ablation_lambda}, the introductions of additional supervise loss terms can help improve the segmentation performance. And the best results is achieved with $\lambda=1.0$.   

\begin{table}[t] \small
	\vspace{-5mm}
	\centering
	\caption{Impact of $\lambda$ in auxiliary loss.}
	\label{tab:ablation_lambda}
	\renewcommand{\tabcolsep}{4.0mm}
	\resizebox{0.9\linewidth}{!}{%
		\begin{tabular}{c|cccc}
			\hline
			$\lambda$ & 0 & 0.1 & 0.5 & 1.0 \\
			\hline
			mIoU & 59.2 & 61.1 & 62.8 & 64.3 \\
			\hline
	\end{tabular}}
	\vspace{-5mm}
\end{table}

\begin{table}[t]\footnotesize
	\centering
	\caption{Ablation studies for auxiliary loss module on different backbone.} \label{tab:abla_backbone}
	\begin{tabular}{c|c|c}
		\hline
		Backbones &Params(M) &mIOU\\
		\hline 
		HarDNet(vinilla) & 2.735 & 55.1\\	\hline
		HarDNet(ours) &3.138 &  58.7 \\	\hline
		HarDNet(ours) + Plan B & 3.146 & 61.3 \\	\hline
	\end{tabular}
	\vspace{-5mm}
\end{table}

\noindent\textbf{Effectiveness on Different Backbone.} In Table~\ref{tab:abla_backbone}, we use a different feature extraction backbone, \textbf{HarDNet} (Harmonic DenseNet), to verify the generalization ability of the auxiliary loss modules. Compared with ResNet and DenseNet, HarDNet can achieve comparable accuracy in several tasks while significantly reducing GPU running time. First, we use FC-HarDNet-70 to conduct experiment, and obtain 55.1\% mIoU. Then, we carefully optimize the structure of HarDNet to construct a more powerful baseline with addition of fewer number of parameters. Finally, we integrate the auxiliary losses, and achieve a consistent model performance improvement. 

\section{Conclusion}
In this paper, we propose a newly-designed networks,termed CENet, for LiDAR point cloud segmentation task. It is a concise and efficient model. Based on analysis of previous studies, we choose to use standard convolution with larger kernel size, and the SiLU as well as Hardswish activation functions, to improve learning capability of our network. Then, we embed multiple auxiliary segmentation heads to further improve the power of learned features without introduction of parameters and efficiency cost. Experimental results on SemanticKITTI and SemanticPOSS demonstrate that our CENet can achieve state-of-the-art performance. 


\small 
\bibliographystyle{IEEEbib}
\bibliography{icme2022template}

\begin{thebibliography}{10}

\bibitem{behley2019semantickitti}
Jens Behley, Martin Garbade, Andres Milioto, Jan Quenzel, Sven Behnke, Cyrill
  Stachniss, and Jurgen Gall,
\newblock ``Semantickitti: A dataset for semantic scene understanding of lidar
  sequences,''
\newblock in {\em ICCV}, 2019, pp. 9297--9307.

\bibitem{feng2020deep}
D.~Feng, C.~Haase-Sch{\"u}tz, L.~Rosenbaum, H.~Hertlein, C.~Glaeser, F.~Timm,
  W.~Wiesbeck, and K.~Dietmayer,
\newblock ``Deep multi-modal object detection and semantic segmentation for
  autonomous driving: Datasets, methods, and challenges,''
\newblock {\em {IEEE} Trans. Intell. Transp. Syst.}, 2020.

\bibitem{pan2020semanticposs}
Yancheng Pan, Biao Gao, Jilin Mei, Sibo Geng, Chengkun Li, and Huijing Zhao,
\newblock ``Semanticposs: A point cloud dataset with large quantity of dynamic
  instances,''
\newblock in {\em {IV}}. IEEE, 2020, pp. 687--693.

\bibitem{qi2017pointnet++}
Charles~Ruizhongtai Qi, Li~Yi, Hao Su, and Leonidas~J Guibas,
\newblock ``Pointnet++: Deep hierarchical feature learning on point sets in a
  metric space,''
\newblock {\em NIPS}, vol. 30, 2017.

\bibitem{thomas2019kpconv}
Hugues Thomas, Charles~R Qi, Jean-Emmanuel Deschaud, Beatriz Marcotegui,
  Fran{\c{c}}ois Goulette, and Leonidas~J Guibas,
\newblock ``Kpconv: Flexible and deformable convolution for point clouds,''
\newblock in {\em ICCV}, 2019, pp. 6411--6420.

\bibitem{hu2020randla}
Qingyong Hu, Bo~Yang, Linhai Xie, Stefano Rosa, Yulan Guo, Zhihua Wang, Niki
  Trigoni, and Andrew Markham,
\newblock ``Randla-net: Efficient semantic segmentation of large-scale point
  clouds,''
\newblock in {\em CVPR}, 2020, pp. 11108--11117.

\bibitem{tang2020searching}
Haotian Tang, Zhijian Liu, Shengyu Zhao, Yujun Lin, Ji~Lin, Hanrui Wang, and
  Song Han,
\newblock ``Searching efficient 3d architectures with sparse point-voxel
  convolution,''
\newblock in {\em ECCV}. Springer, 2020, pp. 685--702.

\bibitem{zhou2020cylinder3d}
H.~Zhou, X.~Zhu, X.~Song, Y.~Ma, Z.~Wang, H.~Li, and D.~Lin,
\newblock ``Cylinder3d: An effective 3d framework for driving-scene lidar
  semantic segmentation,''
\newblock {\em arXiv preprint arXiv:2008.01550}, 2020.

\bibitem{wu2018squeezeseg}
Bichen Wu, Alvin Wan, Xiangyu Yue, and Kurt Keutzer,
\newblock ``Squeezeseg: Convolutional neural nets with recurrent crf for
  real-time road-object segmentation from 3d lidar point cloud,''
\newblock in {\em ICRA}. IEEE, 2018, pp. 1887--1893.

\bibitem{milioto2019rangenet++}
Andres Milioto, Ignacio Vizzo, Jens Behley, and Cyrill Stachniss,
\newblock ``Rangenet++: Fast and accurate lidar semantic segmentation,''
\newblock in {\em IROS}. 2019, pp. 4213--4220, IEEE.

\bibitem{fan2021rangedet}
Lue Fan, Xuan Xiong, Feng Wang, Naiyan Wang, and Zhaoxiang Zhang,
\newblock ``Rangedet: In defense of range view for lidar-based 3d object
  detection,''
\newblock in {\em ICCV}, 2021, pp. 2918--2927.

\bibitem{guo2020deep}
Y.~Guo, H.~Wang, Q.~Hu, H.~Liu, L.~Liu, and M.~Bennamoun,
\newblock ``Deep learning for 3d point clouds: A survey,''
\newblock {\em {IEEE} Trans. Pattern Anal. Mach. Intell.}, 2020.

\bibitem{kochanov2020kprnet}
Deyvid Kochanov, Fatemeh~Karimi Nejadasl, and Olaf Booij,
\newblock ``Kprnet: Improving projection-based lidar semantic segmentation,''
\newblock {\em arXiv preprint arXiv:2007.12668}, 2020.

\bibitem{Lite-HDSeg}
Ryan Razani, Ran Cheng, Ehsan Taghavi, and Bingbing Liu,
\newblock ``Lite-hdseg: Lidar semantic segmentation using lite harmonic dense
  convolutions,''
\newblock in {\em {ICRA}}. 2021, pp. 9550--9556, {IEEE}.

\bibitem{zhao2021fidnet}
Yiming Zhao, Lin Bai, and Xinming Huang,
\newblock ``Fidnet: Lidar point cloud semantic segmentation with fully
  interpolation decoding,''
\newblock in {\em {IROS}}. 2021, pp. 4453--4458, {IEEE}.

\bibitem{qiu2021semantic}
Shi Qiu, Saeed Anwar, and Nick Barnes,
\newblock ``Semantic segmentation for real point cloud scenes via bilateral
  augmentation and adaptive fusion,''
\newblock in {\em {CVPR}}, 2021, pp. 1757--1767.

\bibitem{qi2017pointnet}
Charles~R Qi, Hao Su, Kaichun Mo, and Leonidas~J Guibas,
\newblock ``Pointnet: Deep learning on point sets for 3d classification and
  segmentation,''
\newblock in {\em CVPR}, 2017, pp. 652--660.

\bibitem{graham20183d}
Benjamin Graham, Martin Engelcke, and Laurens Van Der~Maaten,
\newblock ``3d semantic segmentation with submanifold sparse convolutional
  networks,''
\newblock in {\em CVPR}, 2018, pp. 9224--9232.

\bibitem{wu2019squeezesegv2}
Bichen Wu, Xuanyu Zhou, Sicheng Zhao, Xiangyu Yue, and Kurt Keutzer,
\newblock ``Squeezesegv2: Improved model structure and unsupervised domain
  adaptation for road-object segmentation from a lidar point cloud,''
\newblock in {\em ICRA}. IEEE, 2019, pp. 4376--4382.

\bibitem{xu2020squeezesegv3}
Chenfeng Xu, Bichen Wu, Zining Wang, Wei Zhan, Peter Vajda, Kurt Keutzer, and
  Masayoshi Tomizuka,
\newblock ``Squeezesegv3: Spatially-adaptive convolution for efficient
  point-cloud segmentation,''
\newblock in {\em ECCV}. Springer, 2020, pp. 1--19.

\bibitem{ding2021repvgg}
Xiaohan Ding, Xiangyu Zhang, Ningning Ma, Jungong Han, Guiguang Ding, and Jian
  Sun,
\newblock ``Repvgg: Making vgg-style convnets great again,''
\newblock in {\em CVPR}, 2021, pp. 13733--13742.

\bibitem{cortinhal2020salsanext}
Tiago Cortinhal, George Tzelepis, and Eren~Erdal Aksoy,
\newblock ``Salsanext: Fast, uncertainty-aware semantic segmentation of lidar
  point clouds,''
\newblock in {\em ISVC}. Springer, 2020, pp. 207--222.

\bibitem{alnaggar2021multi}
Yara~Ali Alnaggar, Mohamed Afifi, Karim Amer, and Mohamed ElHelw,
\newblock ``Multi projection fusion for real-time semantic segmentation of 3d
  lidar point clouds,''
\newblock in {\em WACV}, 2021, pp. 1800--1809.

\bibitem{li2020multi}
S.~Li, X.~Chen, Y.~Liu, D.~Dai, C.~Stachniss, and J.~Gall,
\newblock ``Multi-scale interaction for real-time lidar data segmentation on an
  embedded platform,''
\newblock {\em {IEEE} Robotics Autom. Lett.}, 2021.

\end{thebibliography}

\end{document}